\documentclass[9pt,final,a4paper]{IEEEtran}
\IEEEoverridecommandlockouts
\usepackage{cite}
\usepackage{amsmath,amssymb,amsfonts}
\usepackage[algoruled]{algorithm2e}
\usepackage{graphicx}
\usepackage{textcomp}
\usepackage{xcolor}
\usepackage{subcaption}
    
\begin{document}

\title{A Novel Random Forest Dissimilarity Measure for Multi-View Learning}
\author{
    \IEEEauthorblockN{Hongliu Cao\IEEEauthorrefmark{1}\IEEEauthorrefmark{2}, Simon Bernard\IEEEauthorrefmark{1}, Robert Sabourin\IEEEauthorrefmark{2}, Laurent Heutte\IEEEauthorrefmark{1}}
    
    \IEEEauthorblockA{\IEEEauthorrefmark{1}LITIS, Université de Rouen Normandie, 76000 Rouen, France \\ caohongliu@gmail.com, simon.bernard@univ-rouen.fr, laurent.heutte@univ-rouen.fr}
    
    \IEEEauthorblockA{\IEEEauthorrefmark{2}LIVIA, École de Technologie Supérieure (ÉTS), Université du Québec, Montreal, QC, Canada \\ robert.sabourin@etsmtl.ca}
}

\maketitle

\begin{abstract}
Multi-view learning is a learning task in which data is described by several concurrent representations. Its main challenge is most often to exploit the complementarities between these representations to help solve a classification/regression task. This is a challenge that can be met nowadays if there is a large amount of data available for learning. However, this is not necessarily true for all real-world problems, where data are sometimes scarce (e.g. problems related to the medical environment). In these situations, an effective strategy is to use intermediate representations based on the dissimilarities between instances. This work presents new ways of constructing these dissimilarity representations, learning them from data with Random Forest classifiers. More precisely, two methods are proposed, which modify the Random Forest proximity measure, to adapt it to the context of High Dimension Low Sample Size (HDLSS) multi-view classification problems. The second method, based on an Instance Hardness measurement, is significantly more accurate than other state-of-the-art measurements including the original RF Proximity measurement and the Large Margin Nearest Neighbor (LMNN) metric learning measurement.
\end{abstract}


\section{Introduction}

Multi-view learning (MVL) is a machine learning task for which the data is described by several concurrent representations. Nowadays, many real-world learning problems are naturally multi-view in the sense that the instances are supposed to be too complex to be described by a single numerical representation. One of the early examples is given in \cite{Blum1998}, where Web-pages are represented by their textual content (first view) and by the anchor texts attached to hyperlinks pointing to them (second view). More recent examples are audio and video captures of a same event or sets of images accompanied by textual data \cite{Zhao2017, Li2019}.

The emergence of MVL techniques stems from the realization that concatenating all the features/modalities to form one single representation and applying traditional learning algorithms often leads to over-fitting problems when the number of instances is relatively small compared to the number of features. In contrast, learning from each view separately and trying to maximize the agreement afterwards help to overcome this over-fitting problem while allowing at the same time to take the specific statistical properties of each view into account \cite{Zhao2017}. Moreover, the rationale for MVL is that the views contain complementary information and the challenge is to exploit this to better solve the learning task \cite{Li2019}.

The most popular approach in multi-view learning usually consists in learning separate models on each view and in combining these models, often by jointly adjusting them in order to maximize their agreement \cite{Zhao2017}. The most representative methods of the kind are Co-training methods \cite{Blum1998, Zhao2017}. However, this requires to use additional -- often unsupervised -- data for the combination step, which is likely to be impossible for many real-world problems for which data are particularly difficult to collect. Typical examples are machine learning problems related to the medical field where sample sizes are particularly small \cite{Cao2019}. In such cases, the so-called High Dimension, Low Sample Size (HDLSS) problems, method like Co-training techniques are not applicable.

In \cite{Cao2019}, a different strategy is used for HDLSS multi-view learning, based on learning intermediate representations from each view and on combining them in order to form a new joint representation from which the model is learnt. In this framework, the intermediate representations are dissimilarity representations, that is to say description spaces in which each instance is described through its dissimilarities to all the training instances. This has three main advantages for HDLSS multi-view learning: i) these representations doesn't require a large amount of data for learning, ii) the dissimilarity spaces dimensions are much lower than the original views dimensions, and iii) the combination step is much more straightforward and versatile \cite{Cao2019}.

The originality of the method in \cite{Cao2019} is the use of Random Forest (RF) classifiers to learn the dissimilarity representations. RF are powerful and versatile classifiers that incorporate a measure of (dis)similarity, with good theoretical properties \cite{Davie2014, Scorn2016} and that can be learned in a non-parametric way, i.e. without prior formulation of the measure. The objective of the present work is to deepen the use of RF classifiers for learning dissimilarity representations in the context of HDLSS multi-view learning. We propose novel ways to learn dissimilarities from RF classifiers, in order to build better intermediate representations for MVL classification tasks.

The reminder of the paper is organized as follows: Section \ref{sec:rfdmvl} gives the general Random Forest Dissimilarity framework for Multi-view learning; Section \ref{sec:dissrep} explains the way intermediate dissimilarity representations are built in this framework; Section \ref{sec:rfdissrep} describes the Random Forest Dissimilarity measure and gives our two new variants; and Section \ref{sec:xp} details the experimental validation and analysis of the results.

\section{Random Forest Dissimilarity for multi-view learning}
\label{sec:rfdmvl}

In a traditional supervised learning task, each instance is described by a single vector of $m$ features. For MVL tasks, each instance is described by $Q$ different vectors. Therefore, the task is to learn a model $h$:
\begin{equation}
	h : \mathcal{X}^{(1)} \times \mathcal{X}^{(2)} \times \dots \times \mathcal{X}^{(Q)} \rightarrow \mathcal{Y}
\end{equation}
where the $\mathcal{X}^{(q)}$ are the $Q$ input domains, i.e. the views. These views are generally very heterogeneous, of different nature and of different dimensions $m_1$ to $m_Q$. Therefore, for this type of learning task, the training set $T$ is actually composed of $Q$ training subsets noted :
\begin{equation}
	T^{(q)} = \left\lbrace (\mathbf{x}^{(q)}_1,y_1), (\mathbf{x}^{(q)}_2,y_2),\dots,(\mathbf{x}^{(q)}_n,y_n) \right\rbrace, \forall q=1..Q
\end{equation}

The Random Forest Dissimilarity (RFD) framework consists first in building dissimilarity representations from each of the $T^{(q)}$. As explained in detail in a recent review on dissimilarity based pattern recognition \cite{Costa2019}, one of the most widely known and used dissimilarity strategies is the so-called \textit{dissimilarity space} approach. It consists in building a dissimilarity matrix from the training set $T$ and in learning a model from this matrix. In the present work, a dissimilarity matrix is a $n \times n$ matrix, built from the $n$ training instances such as:
\begin{equation}
    \mathbf{D}(T,T) = \begin{bmatrix}
        d(\mathbf{x}_1, \mathbf{x}_1) & d(\mathbf{x}_1, \mathbf{x}_2) & \dots & d(\mathbf{x}_1, \mathbf{x}_n)\\
        d(\mathbf{x}_2, \mathbf{x}_1) & d(\mathbf{x}_2, \mathbf{x}_2) & \dots & d(\mathbf{x}_2, \mathbf{x}_n)\\
        \dots & \dots & \dots & \dots \\
        d(\mathbf{x}_n, \mathbf{x}_1) & d(\mathbf{x}_n, \mathbf{x}_2) & \dots & d(\mathbf{x}_n, \mathbf{x}_n)\\
     \end{bmatrix}
\end{equation}
where $d$ stands for a dissimilarity measure and $\mathbf{x}_i$ are the training instances.

Once these matrices are built from each of the $Q$ views, they have to be merged in order to build the joint dissimilarity matrix $\mathbf{D}_H$ that will serve as a new training set for an additional learning phase. This additional learning phase can be realized with any learning algorithm, since the goal is to address the classification task. For simplicity and because they are as accurate as they are versatile, the same Random Forest method used to calculate the dissimilarities is also used in this final learning stage.

As for the merging step, it can be straightforwardly done by a simple average of the $Q$ RFD matrices:
\begin{equation}
    \mathbf{D}_H = \frac{1}{Q}\sum_{q=1}^{Q} \mathbf{D}_H^{(q)}
\end{equation}
The whole RFD based MVL procedure is summarized in Algorithm \ref{algo:rfdis}.

\begin{algorithm}[ht]
    \small
    \SetAlgoLined
    \LinesNumbered
    \DontPrintSemicolon
    \KwIn{$T^{(q)}$, $\forall q=1..Q$: the $Q$ training sets, composed of $n$ instances}
    \KwIn{$RF(.)$: The Breiman's RF learning procedure}
    \KwIn{$RFD(.,.|.)$: the $RFD$ measure}
    \KwOut{$H^{(q)}$: $Q$ RF classifiers}
    \KwOut{$H_{final}$: the final RF classifier}
    \BlankLine
    \For{$q=1..Q$}{
        $H^{(q)} = RF(T^{(q)})$\;
        \tcp{Build the $n \times n$ RFD matrix $\mathbf{D}_H^{(q)}$:}
        \ForAll{$\mathbf{x}_i \in T^{(q)}$}{
            \ForAll{$\mathbf{x}_j \in T^{(q)}$}{
                $\mathbf{D}_H^{(q)}[i,j] = RFDis(\mathbf{x}_i,\mathbf{x}_j | H^{(q)})$\;
            }
        }
    }
    \tcp{Build the $n \times n$ average RFD matrix $\mathbf{D}_H$:}
    $\mathbf{D}_H = \frac{1}{Q}\sum_{q=1}^{Q} \mathbf{D}_H^{(q)}$\;
    \tcp{Train the final classifier on $\mathbf{D}_H$:}
    $H_{final} = RF(\mathbf{D}_H)$\;
    \caption{The RFD multi-view learning procedure\label{algo:rfdis}}
\end{algorithm}

\section{Learning dissimilarity representations}
\label{sec:dissrep}

The main challenge in using dissimilarity for MVL, is to construct the most relevant dissimilarity representations from each of the $Q$ views. These representations must best reflect the specificities of each view in order to exploit their complementarities later on. We argue that the best way to do so, is to learn these representations from each of the $T^{(q)}$ separately in a supervised way, i.e. by taking the outputs $y_i, i=1..n$ into consideration. The reason is that, in multi-view learning, the views are likely to contribute in different ways to the final task, and therefore, we want the dissimilarity representation to reflect these contributions the best possible.

A first approach to learn the dissimilarity matrices introduced in the previous section would be to use a \textit{metric learning} methods \cite{Bellet2015}, which purpose is to learn a distance metric function $d(\mathbf{x}_i, \mathbf{x}_j)$, for all ($\mathbf{x}_i,\mathbf{x}_j) \in T \times T$, and to use this function to calculate each element of the matrices. A metric learning algorithm basically aims at finding the parameters of the metric (e.g. the Minkowski or Mahalanobis distances) such that it best agrees with some ground truth constraints. In a fully supervised learning context, these constraints (e.g. \textit{must-link/cannot-link} constraints) are usually inferred from the training instances based on a notion of neighborhood \cite{Bellet2015}. Nevertheless, one of the main difficulties in using metric learning methods in our case, is that it usually requires to learn a $m \times m$ matrix of parameters, $m$ being the dimension of the initial description space. This is a significant hurdle for problems that have more features than training instances, as in HDLSS multi-view learning.

Another approach is to use \textit{random partitions} \cite{Davie2014}. Random partitions adopt a different approach in the sense that the method strives to infer the model from the training instances only, without any prior formulation of the measure or any similarity constraints. The key idea of random partitions is to define multiple randomized partitions of the input space in such a way it forms homogeneous groups (clusters) of instances. It has been proven that such random partitions can be used to define kernels, which can be viewed as a (dis)similarity measurement \cite{Davie2014, Scorn2016}. Beyond this mathematical demonstrations, random partitions can be directly used in practice to measure similarities, as with the well-known proximity measurement of random forests \cite{Breim2001, Verik2011, Cao2019}. The principle is to estimate the similarity between two instances by the number of times these instances are grouped in the same cluster, over all the partitions. This is the approach we adopt in the present work for two main reasons:
\begin{itemize}
    \item These approaches are a lot more versatile than metric learning methods, since they do not require to choose a generic distance metric function to optimize beforehand, neither to infer similarity constraints from the training instances.
    \item Depending on the method used to build the random partitions, it is potentially a lot more robust to high dimensions than metric learning methods.
\end{itemize}
In \cite{Cao2019}, this approach has been successfully used for dissimilarity based multi-view classification by using random forest classifiers for building the random partitions. RF were chosen here precisely because of their robustness to high dimensions, and also because they allow to exploit the class membership for learning the dissimilarities. The following section provides a full explanation of how to do this.

\section{Random Forest Dissimilarity representations}
\label{sec:rfdissrep}

\begin{figure*}[htbp]
    \centering
    \begin{subfigure}[t]{0.5\linewidth}
        \centerline{\includegraphics[width=\linewidth]{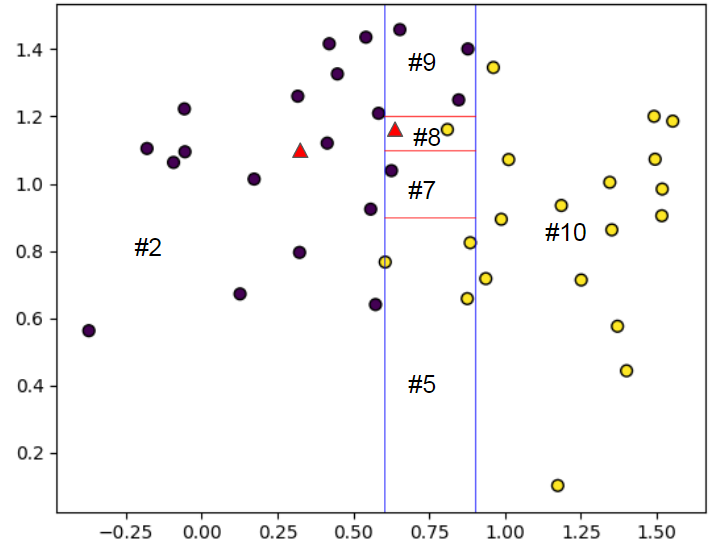}}
        \caption{}
        \label{fig:treeExplea}
    \end{subfigure}
    ~
    \begin{subfigure}[t]{0.3\linewidth}
        \centerline{\includegraphics[width=.9\linewidth]{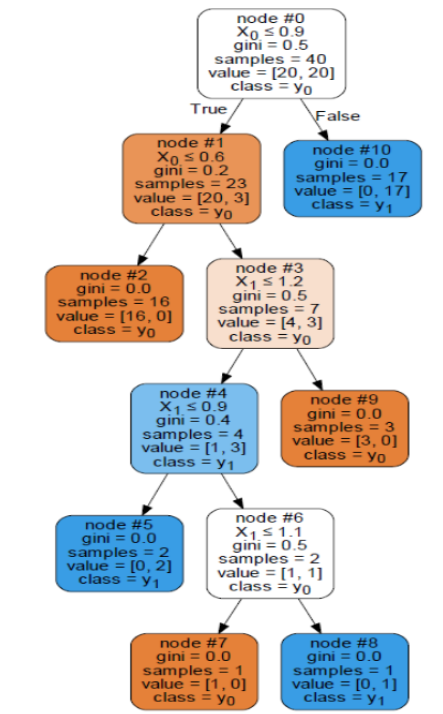}}
        \caption{}
        \label{fig:treeExpleb}
    \end{subfigure}
    \caption{Illustration of a partition (a) given by a decision tree (b) on a synthetic dataset.}
    \label{fig:treeExple}
\end{figure*}

In this work, the name "Random Forest" (RF) refers to the Breiman's reference method \cite{Breim2001}. Let us briefly recall its procedure to build a forest of $P$ decision trees, from a training set $T$. First, a bootstrap sample is built by the random drawing with replacement of $n$ instances, amongst the $n$ training instances available in $T$. Each of these bootstrap samples is then used to build one tree. During this induction phase, at each node of the tree, a splitting rule is designed by selecting a feature over $mtry$ features randomly drawn from the $m$ available features. The feature retained for this splitting rule is the one among the $mtry$ that maximizes the splitting criterion. At last, the trees in RF classifiers are grown to their maximum depth, that is to say when all their terminal nodes (the leaves) are pure. In a given tree, these terminal nodes altogether form one random partitions, since they divide the input space into several area in which the instances are supposed to belong to the same class. An illustration is given in Figure \ref{fig:treeExple}.

For predicting the class of a given instance $\mathbf{x}$ with a decision tree, $\mathbf{x}$ goes down the tree structure from its root to one of its leaves. The descending path followed by $\mathbf{x}$ is determined by successive tests on the values of its features, one per node along the path. The prediction is given by the leaf in which $\mathbf{x}$ has landed. The key point here is that, if two instances land in the same terminal node, they are likely to belong to the same class and they are also likely to share similarities in their feature vectors, since they have followed the same descending path. This is the main motivation behind using RF for measuring dissimilarities between instances.

Let us formally define the dissimilarity measure $d_p$, obtained from a decision tree $h_p$: let $\mathcal{L}_p$ denote the set of leaves of $h_p$, and let $l_p(\mathbf{x})$ denote a function from the input domain $\mathcal{X}$ to $\mathcal{L}_p$, that returns the leaf of $h_p$ where $\mathbf{x}$ lands when one wants to predict its class. The dissimilarity measure $d_p$ is defined as in Equation \ref{eq:dissDT}: if two training instances $\mathbf{x}_i$ and $\mathbf{x}_j$ land in the same leaf of $h_p$, then the dissimilarity between both instances is set to $0$, else it is equal to $1$.
\begin{equation}
    \label{eq:dissDT}
    d_p(\mathbf{x}_i, \mathbf{x}_j) = 
    \left\{ 
        \begin{array}{ll}
            0, & if \  l_p(\mathbf{x}_i)=l_p(\mathbf{x}_j) \\
            1, & otherwise
        \end{array}
    \right.
\end{equation}
Now, the measure $d_H(\mathbf{x}_i, \mathbf{x}_j)$ derived from the whole forest consists in calculating $d_p$ for each tree in the forest, and in averaging the resulting dissimilarity values over the $P$ trees, as follows:
\begin{equation}
    \label{eq:dissRF}
    d_H(\mathbf{x}_i, \mathbf{x}_j) = \frac{1}{P} \sum_{p=1}^{P} d_p(\mathbf{x}_i, \mathbf{x}_j)
\end{equation}
Similarly to the way the predictions are given by a forest, the rationale is that the accuracy of the dissimilarity measure $d_H$ relies essentially on the averaging over a large number of trees. Note that this measure is the opposite measure of the more widely used RF proximity measure, denoted $s_H$ (or $s_p$) in the following.

Once a RF is learnt from $T$, building the dissimilarity matrix is quite straightforward: it simply consist in using $d_H$ to calculate the dissimilarities between each pair of training instances. \\

Whereas the measure explained above has been widely used for different machine learning tasks (e.g. \cite{Shi2012, Gray2013}), very few works have focus on the method itself. However, we believe that the measurement as it is calculated here is rather rough and should be further refined, particularly in the context of HDLSS multi-view learning. The same idea is shared in \cite{Englu2012}, where the authors state that the similarity values provided by any tree $h_p$ (i.e. $s_p(\mathbf{x}_i,\mathbf{x}_j)$) is a too simple binary measure that could lead to inaccurate measurement $d_H(\mathbf{x}_i,\mathbf{x}_j)$ in case the forest is composed of too few trees. The solution they propose is to estimate $s_p(\mathbf{x}_i,\mathbf{x}_j)$ by taking into account the length of the path that separate the two leaves in which $x_i$ and $x_j$ has landed. Assume $\mathbf{x}_i$ ends in a leaf node $n_i$ and $\mathbf{x}_j$ ends in another leaf node $n_j$. The distance $g_{ijp}$ between $n_i$ and $n_j$ from $h_p$ is the number of edges that composed the path from $n_i$ to $n_j$. The similarity between instances in $n_i$ and $n_j$ is then obtained with:
\begin{equation}
    \label{dis1}
    s_p(x_i,x_j) = \frac{1}{e^{w.g_{ijp}}}
\end{equation}
where $w$ is a hyper-parameter that controls the influence of the $g_{ijp}$. For example, if one consider the two instances represented by red triangles in Figure \ref{fig:treeExplea}, $g_{ijp} = 5$, from node $\#2$ to node $\#8$.

However, this method presents several limitations from our point of view. Firstly, the range and the precision of the $s_p(\mathbf{x}_i,\mathbf{x}_j)$ values depend on the size of the tree, which is strongly problem-dependent. Secondly, the proposed method is controlled by a new hyper-parameter, potentially difficult to tune in the HDLSS setting due to the lack of validation data. Thirdly and most importantly, the path length $g_{ijp}$ does not reflect the (dis)similarity between instances from our point of view. For example, consider the smallest non-zero value $g_{ijp} = 2$, i.e. for two leaves that share the same parent node. The instances are likely to share similarities in their description vectors but they are also likely to belong to different classes, since the role of a splitting node is to further separate the classes. Another pathological case is illustrated by the 'red triangle' instances in Figure \ref{fig:treeExple}. From this tree, $g_{\#2\#8}=5$ and $g_{\#2\#10}=3$, which implies that the left red triangle is considered more similar to any points in the right side of Figure \ref{fig:treeExplea} (node $\#10$) than it is to the second red triangle (node $\#8$), which is obviously not true. 

In the following section, we present two novel methods for measuring dissimilarities with RF classifiers, that follow the same goal of proposing a more accurate measurement but that overcome the aforementioned limitations and that also better suits to the HDLSS multi-view learning tasks.

\subsection{RFD based on node confidence}
\label{ssec:rfdnc}

The first method we propose aim at computing the dissimilarity between two instances from a confidence estimate on the leaves of the trees. The motivation behind this can be easily illustrated from Figure \ref{fig:treeExplea}. When we consider the two 'red triangle' instances, we can see that the one in node $\#2$ will be considered similar to all the 'blue circle' instances in the same area, which seems correct. On the other hand, we can see that the 'red triangle' instance in node $\#8$ will be considered similar only to the 'yellow circle' instance in the same node, which is questionable. This is mainly due to the fact that both nodes are not as reliable as each others: instances are more likely to be wrongly predicted in node $\#8$ than in node $\#2$. We thus propose to weight the dissimilarity measure with an estimate of the confidence given by the leaves. 

For estimating this confidence, we propose to use the well-known Out-Of-Bag (OOB) mechanism. When the Bagging principle is used for building a RF, a given bootstrap sample $T_p$ is constructed by random draw with replacement from $T$. In that case, there may be training instances from $T$ that are not included in $T_p$ and consequently, that have not been used for learning the tree $h_p$. These instances are called the out-of-bag instances of $h_p$ \cite{Breim2001}. We propose to compute a confidence estimate for each leaf of $h_p$ by using its OOB instances. More precisely, the confidence score of a leaf from $h_p$ is estimated by the ratio of the instances in this leaf that have been correctly predicted. Hence, the weight associated to $h_p$ for any instance $\mathbf{x}_t$ can be defined as:
\begin{equation}
    \label{prop1}
    w_p (\mathbf{x}_t) =  \frac{1}{| l_p(\mathbf{x}_t) |} \sum_{\mathbf{x}_i\in l_p(\mathbf{x}_t)}I(h_p(\mathbf{x}_i) = y_i)
\end{equation}
where $| l_p(\mathbf{x}_t) |$ is the number of training instances, including the OOB instances, that have landed in the same terminal node as $\mathbf{x}_t$. Such an estimate would ideally be computed from an independent validation set, but let us recall that the HDLSS setting makes often impossible to obtain additional instances for that purpose. So here, we hope for the OOB mechanism to help identifying the unreliable leaf nodes like the node $\#8$ in Figure \ref{fig:treeExple} without the need for an independent validation set and also without only relying on the instances used for building the trees.

This method is quite straightforward, but it still has an important limitation from our point of view: for building the dissimilarity representation, a given instance will have the same dissimilarity value to all the training instances of the node in which it is located. For example, the 'red triangle' instance in node $\#2$ in Figure \ref{fig:treeExplea} will have the same dissimilarity value to all the 'blue circle' instances in the same node, from the first 'blue circle' point in the very left to the ones the closest to the right bound of node $\#2$. It is desirable to go even further in the refinement of the dissimilarity measure, by making it possible to differentiate instances within the same node. This is the motivation behind the second approach we propose in this work, detailed in the following section. 

\subsection{RFD based on instance hardness}
\label{ssec:rfdih}

The node confidence estimator in the previous method is an indicator of the intrinsic difficulty of classifying any instance in a given leaf, i.e. at a node level. With this second method that we propose, we wish to do the same but at the scale of each instance separately within the leaf, i.e. at an instance level. For that purpose, we propose to use an Instance Hardness (IH) measure. In the literature, there exists many different IH measures for analyzing the data complexity at an instance level, most of them being detailed in \cite{Smith2014}. According to the analysis given in this article, the most appropriate and relevant IH measure to achieve our goal is the k-Disagreeing Neighbors (kDN) measure. This measure is the ratio of instances in the neighborhood of a given instance that belong to a different class:
\begin{equation}
    \label{eq:kdn}
    kDN (\mathbf{x}_i) = \frac{|\mathbf{x}_j: \mathbf{x}_j \in kNN(\mathbf{x}_i) \cap y_j \neq y_i|}{k}
\end{equation}
where $kNN(\mathbf{x}_i)$ stands for the $k$ nearest neighbors of $\mathbf{x}_i$, e.g. according to the Euclidean distance, and where $y_i$ (resp. $y_j$) is the true class of $\mathbf{x}_i$ (resp. $\mathbf{x}_j$). The kDN measure is quite straightforward and easy to understand. If a given instance is mainly surrounded by instances of the same class, the kDN value is close to 0 and the instance can be considered easy to classify. On the other hand, if its nearest neighbors are all from a different class, the kDN value is close to 1 and the instance is considered hard to classify. This measure has been successfully used in several recent works due to its simplicity and interpretability (e.g. in \cite{Cruz2017}).

In our method, the kDN measure is used to weight the dissimilarity values $d_p(\mathbf{x},\mathbf{x}_i)$, for any $\mathbf{x}$ and for $\mathbf{x}_i \in T$. More precisely, $d_p(\mathbf{x},\mathbf{x}_i)$ is computed as follows:
\begin{equation}
    \label{eq:dissDT_IH}
    d_p(\mathbf{x}, \mathbf{x}_i) = 
    \left\{ 
        \begin{array}{ll}
            kDN(\mathbf{x}_i), & if \  l_p(\mathbf{x})=l_p(\mathbf{x}_i) \\
            1, & otherwise
        \end{array}
    \right.
\end{equation}

However, it is not relevant in our case to compute the kDN values globally, that is to say in the whole feature space. The first reason is that the kDN measure is quite sensitive to high dimensions due to the use of the Euclidean distance measure to define the neighborhood. The Euclidean distance is known to suffer from the curse of dimensionality \cite{Angiu2017, Feldb2019}. The second and most important reason is that our goal is to determine whether a tree is reliable for measuring the dissimilarity to a given training instance, taking into account the leaf in which it is located. For this, we can rely only on the features used by the decision path leading to this leaf. Indeed, it is likely that this path exploits only a subset of the features, potentially small for HDLSS problems. For example, a binary decision tree built on a training set with $n$ instances will have a maximum of $n$ leaf nodes and $n-1$ split nodes (corresponding to the worst case where all the leaves contain one instance each). As a consequence, there are a maximum of $n-1$ features that are used in any decision path. For classification tasks, the number of leaves in a tree is usually much smaller than $n$ because leaf nodes are likely to contain more than one instance. Hence, for HDLSS problem where $n$ is much smaller than the total number of features $m$, only a small subset of features is used for determining the leaf of any instance.

\begin{figure}[htbp]
    \centerline{\includegraphics[width=.95\linewidth]{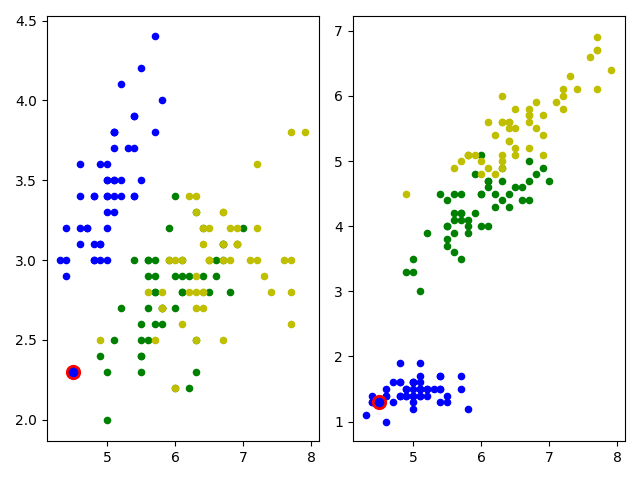}}
    \caption{Scatter plot of the Iris dataset into two of its 2D subspaces. The red circled blue dot is the same point projected in both subspaces.}
    \label{fig:iris}
\end{figure}

To further illustrate the interest of measuring kDN in the subspace defined by a leaf path, the Iris toy dataset is plotted in Figure \ref{fig:iris}, with its instances projected in two of its 2D subspaces. The relative position of the red circled instance is very different from one subspace to the other: in the left subplot, it is close to instances from the green class while in the right subplot it is in the core of the blue class cluster. As a consequence, the kDN value computed from the left subspace is likely to be 1 and the kDN value computed from the right subspace is likely to be 0. In our case, this is precisely the kind of phenomenon we want to detect for computing the dissimilarities, in order to reduce the weight of unreliable dissimilarity values in the calculation of the final measurement $d_H$ (Eq. \ref{eq:dissRF}).

Algorithm \ref{algo:rfdih} details the procedure for computing the dissimilarity representation of any given instance $\mathbf{x}$ according to the method we propose, based on the Instance Hardness measure and noted $RFDis_{IH}$ in the following.

\begin{algorithm}[ht]
    \small
    \SetAlgoLined
    \LinesNumbered
    \DontPrintSemicolon
    \KwIn{$T$: a training set of $n$ instances}
    \KwIn{$H$: a RF classifier learned on $T$, with $P$ trees $h_p$}
    \KwIn{$\mathbf{x}$: an instance}
    \KwIn{$k$: number of nearest neighbors for computing $kDN$}
    \KwOut{$\mathbf{d}$: the $RFDis_{IH}$ representation of $x$}
    \BlankLine
    \ForAll{$\mathbf{x}_i \in T$}{
        $\mathbf{s}_i = 0$ \;
        \ForAll{$h_p \in H$}{
            \If{$l_p(\mathbf{x}_i) = l_p(\mathbf{x})$}{
                $kNN(\mathbf{x}_i) = k$ nearest neighbors of $\mathbf{x}_i$ in the subspace formed by $l_p(\mathbf{x}_i)$ \;
                $kDN (\mathbf{x}_i) = \frac{|\mathbf{x}_j: \mathbf{x}_j \in kNN(\mathbf{x}_i) \cap y_j \neq y_i|}{k}$ \;
                \tcp{the similarity between $\mathbf{x}$ and $\mathbf{x}_i$ according to $h_p$:}
                $s_p(\mathbf{x},\mathbf{x}_i) = 1 - kDN(\mathbf{x}_i)$ \;
                $\mathbf{s}_i = \mathbf{s}_i + s_p(\mathbf{x},\mathbf{x}_i)$ \;
            }
        }
        \tcp{the $i$-th value of $\mathbf{d}$ is the average dissimilarity between $\mathbf{x}$ and $\mathbf{x}_i$, over the $P$ trees}
        $\mathbf{d}_i = 1 - (\mathbf{s}_i / P)$ \;
    }
    \caption{Random Forest Dissimilarity with Instance Hardness\label{algo:rfdih}}
\end{algorithm}

\section{Experiments and results}
\label{sec:xp}

The present section details the experiments conducted to compare the two proposed approaches to state-of-the-art methods, on different real-world HDLSS multi-view classification problems.

\subsection{Methods and datasets}

The RFD methods we propose in this work are noted $RFDis_{NC}$ ($NC$ for Node Confidence) and $RFDis_{IH}$ ($IH$ for Instance Hardness) in the following. Both are compared to four methods of measuring dissimilarity:
\begin{enumerate}
    \item the classic Euclidean distance measure noted $EUDis$,
    \item the initial Random Forest Dissimilarity measure described in Section \ref{sec:rfdissrep}, (Eq. \ref{eq:dissDT} and \ref{eq:dissRF}) and noted $RFDis$,
    \item the modified Random Forest Dissimilarity measure from \cite{Englu2012}, presented in section \ref{sec:rfdissrep} and noted $RFDis_{PB}$ and,
    \item following the recommendation of the recent survey \cite{Li2018}, a metric learning method named Large Margin Nearest Neighbors (LMNN) \cite{Domen2005} along with a Principal Component Analysis (PCA). LMNN is one of the most widely-used Mahalanobis distance learning methods. According to the suggestions in \cite{Li2018}, the number of components for PCA is set to 300 and the main parameter of the LMNN method, i.e. the size of neighborhood, is set to 25, except if the dataset presents at least one class with less than 25 instances, in which case it is set to the number of instances that belong to the minority class.
\end{enumerate}
For all the RF classifiers used in this experiment, the number of trees is set to 512 following the conclusions from \cite{Cao2019}, while the other hyper-parameters are set to the default values proposed in the \textit{Scikit-learn} machine learning library \cite{scikit-learn}. All the methods compared in this validation experiment are summed up in Table \ref{tab:methods}.

\begin{table}[htbp]
	\renewcommand{\arraystretch}{1.3}
  \caption{\label{tab:methods} The methods compared in the present experiments}
  \begin{center}
  \begin{tabular}{ll}
    \hline
    \textbf{Method}  & \textbf{Dissimilarity measure}  \\
    \hline
    \hline
    $EUDis$  & Euclidean distance \\ 
    \hline
    $LMNNDis$ & PCA+LMNN \\
    \hline
    $RFDis$ & The reference Random Forest Dissimilarity \\ & (Eq. \ref{eq:dissDT} and \ref{eq:dissRF}) \\
    \hline
    $RFDis_{PB}$  & The $RFDis$ variant from \cite{Englu2012} \\ 
    \hline
    $RFDis_{NC}$ & The method proposed in \\ & Sec. \ref{ssec:rfdnc} (Node Confidence)\\ 
    \hline
    $RFDis_{IH}$ & The method proposed in \\ & Sec. \ref{ssec:rfdih} (Instance Hardness) \\
    \hline
  \end{tabular}
  \end{center}
\end{table}

For this experimental comparison, all the dissimilarity measures from Table \ref{tab:methods} have been used as in the procedure described in Section \ref{sec:rfdmvl} (Algo. \ref{algo:rfdis}). More precisely, each of them has firstly been used to build the dissimilarity matrix from each view; then these dissimilarity matrices have been averaged to form the joint dissimilarity matrix; and finally this joint dissimilarity matrix has been used for the learning of a Random Forest classifier. The two distance measures (Euclidean distance and LMNN) provide unbounded distance values. Therefore, in this experiment, the values have been re-scaled in the interval [0,1] by dividing each dissimilarity vector by its maximum value before the averaging. 

The multi-view datasets used in this experiment are described in Table \ref{tab:data}. All these datasets are publicly available real-world multi-view datasets, supplied with several views of the same instances: \textit{NonIDH1}, \textit{IDHcodel}, \textit{LowGrade} and \textit{Progression} are medical imaging classification problems, with different families of features extracted from different types of radiographic images; \textit{LSVT} and \textit{Metabolomic} are two other medical related classification problems, the first one for Parkinson's disease recognition and the second one for colorectal cancer detection; \textit{BBC} and \textit{BBCSport} are text classification problems from news articles; \textit{Cal7}, \textit{Cal20}, \textit{Mfeat}, \textit{NUS-WIDE2}, \textit{NUS-WIDE3}, \textit{AWA8} and \textit{AWA15} are image classification problems made up with different families of features extracted from the images. More details about these datasets are given in \cite{Cao2019} (and references therein).

\begin{table}[htbp]
	\renewcommand{\arraystretch}{1.1}
    \begin{center}
    \begin{tabular}{|l|c|c|c|c|c|}
      \hline
      & features & instances & views & classes & IR$^{\text a}$ \\
      \hline
      \hline
      AWA8
      & 10940 &640 &6 & 8& 1 \\
      \hline
      AWA15
      & 10940 &1200 & 6 & 15& 1 \\
      \hline
      BBC
      & 13628 &2012 &2 & 5& 1.34 \\
      \hline
      BBCSport
      & 6386 &544 & 2 & 5& 3.16 \\
      \hline
      Cal7
      & 3766 &1474 & 6 & 7& 25.74 \\
      \hline
      Cal20
      & 3766 &2386 & 6 & 20& 24.18 \\
      \hline
      IDHcodel
      & 6746 &67 & 5 & 2&2.94 \\
      \hline
      LowGrade
      & 6746 &75 & 5 & 2&1.4 \\
      \hline
      LSVT
      & 309 &126 & 4 & 2&2 \\
      \hline
      Metabolomic
      & 476 &94 & 3 & 2& 1 \\
      \hline
      Mfeat
      & 649 &600 & 6 & 10&1 \\
      \hline
      NonIDH1
      & 6746 &84 & 5 & 2&3 \\
      \hline
      NUS-WIDE2
      & 639 &442 & 5 & 2& 1.12 \\
      \hline
      NUS-WIDE3
      & 639 &546 & 5 & 3&1.43 \\
      \hline
      Progression
      & 6746 &84 & 5 & 2&1.68 \\
      \hline
    \end{tabular}
    \caption{Real-world multi-view datasets. $^{\text a}$Imbalance Ratio}
    \label{tab:data}
    \end{center}
\end{table}

A stratified random splitting procedure is used and repeated 10 times on each dataset, with 50\% of the instances for training and the remaining 50\% for testing. The mean accuracy, with standard deviations, are computed over the 10 runs and reported in Table \ref{tab:rescomp}, along with the mean rank of each method in the last row. Bold numbers correspond to the best classification results among the five methods on each dataset.

\begin{table*}[htbp]
	\renewcommand{\arraystretch}{1.3}
    \caption{\label{tab:rescomp}Mean classification accuracies (with standard deviation) of the 6 dissimilarity-based multi-view classification methods on 15 real-world multi-view datasets. The last row gives the average rank of each method over the 15 datasets.}	
    \begin{center}
    \begin{tabular}{l c c c c c c} \hline
 & $EUDis$ & $LMNNDis$ & $RFDis$ & $RFDis_{PB} $& $RFDis_{NC}$ & $RFDis_{IH}$ \\ 
 \hline
AWA8 &
 $39.22\% \pm 2.55$ &
 $42.28\% \pm 3.13$ &
 $56.06\% \pm 1.35$ &
 $\mathbf{56.38\% \pm 1.47}$ &
 $56.34\% \pm 1.68$ &
 $56.22\% \pm 1.01$
 \vspace*{0.0mm} \\
AWA15 &
 $24.80\% \pm 0.97$ &
 $28.25\% \pm 1.60$ &
 $37.90\% \pm 1.49$ &
 $37.62\% \pm 1.40$ &
 $37.93\% \pm 1.50$ &
 $\mathbf{38.23\% \pm 0.83}$
 \vspace*{0.0mm} \\
Metabo &
 $\mathbf{69.38\% \pm 2.29}$ &
 $67.08\% \pm 4.04$ &
 $67.71\% \pm 5.12$ &
 $67.50\% \pm 5.76$ &
 $67.08\% \pm 6.31$ &
 $69.17\% \pm 5.80$
 \vspace*{0.0mm} \\
Mfeat &
 $96.00\% \pm 1.45$ &
 $96.87\% \pm 0.79$ &
 $97.56\% \pm 0.99$ &
 $\mathbf{97.63\% \pm 0.95}$ &
 $97.63\% \pm 1.00$ &
 $97.53\% \pm 1.00$
 \vspace*{0.0mm} \\
NUS-WIDE2 &
 $89.52\% \pm 1.44$ &
 $90.33\% \pm 1.55$ &
 $92.49\% \pm 2.01$ &
 $92.49\% \pm 1.81$ &
 $92.67\% \pm 1.47$ &
 $\mathbf{92.82\% \pm 1.93}$
 \vspace*{0.0mm} \\
BBC &
 $85.89\% \pm 1.33$ &
 $93.02\% \pm 1.29$ &
 $92.82\% \pm 0.67$ &
 $93.00\% \pm 0.67$ &
 $92.33\% \pm 0.49$ &
 $\mathbf{95.46\% \pm 0.65}$
 \vspace*{0.0mm} \\
lowGrade &
 $63.72\% \pm 5.12$ &
 $62.33\% \pm 7.04$ &
 $63.48\% \pm 3.76$ &
 $63.72\% \pm 4.67$ &
 $\mathbf{63.95\% \pm 3.64}$ &
 $63.95\% \pm 5.62$
 \vspace*{0.0mm} \\
NUS-WIDE3 &
 $73.92\% \pm 2.40$ &
 $78.02\% \pm 2.69$ &
 $79.41\% \pm 1.94$ &
 $79.64\% \pm 2.19$ &
 $79.91\% \pm 2.14$ &
 $\mathbf{80.32\% \pm 1.95}$
 \vspace*{0.0mm} \\
progression &
 $58.42\% \pm 4.82$ &
 $62.63\% \pm 5.86$ &
 $63.42\% \pm 6.49$ &
 $63.42\% \pm 7.48$ &
 $63.95\% \pm 6.56$ &
 $\mathbf{65.79\% \pm 4.71}$
 \vspace*{0.0mm} \\
LSVT &
 $82.86\% \pm 2.11$ &
 $\mathbf{85.24\% \pm 2.84}$ &
 $83.33\% \pm 3.97$ &
 $82.70\% \pm 3.44$ &
 $83.49\% \pm 3.56$ &
 $84.29\% \pm 3.51$
 \vspace*{0.0mm} \\
IDHCodel &
 $73.53\% \pm 5.42$ &
 $71.47\% \pm 2.30$ &
 $76.47\% \pm 3.95$ &
 $76.47\% \pm 4.16$ &
 $76.18\% \pm 3.82$ &
 $\mathbf{76.76\% \pm 3.59}$
 \vspace*{0.0mm} \\
nonIDH1 &
 $79.07\% \pm 3.45$ &
 $73.26\% \pm 3.49$ &
 $79.53\% \pm 3.57$ &
 $79.53\% \pm 3.72$ &
 $79.77\% \pm 3.46$ &
 $\mathbf{80.70\% \pm 3.76}$
 \vspace*{0.0mm} \\
BBCSport &
 $80.11\% \pm 1.69$ &
 $73.77\% \pm 5.45$ &
 $81.75\% \pm 2.70$ &
 $82.56\% \pm 2.85$ &
 $79.93\% \pm 3.11$ &
 $\mathbf{90.18\% \pm 1.96}$
 \vspace*{0.0mm} \\
Cal20 &
 $84.04\% \pm 0.82$ &
 $87.50\% \pm 0.78$ &
 $89.12\% \pm 0.69$ &
 $89.27\% \pm 1.01$ &
 $89.06\% \pm 1.19$ &
 $\mathbf{89.76\% \pm 0.80}$
 \vspace*{0.0mm} \\
Cal7 &
 $92.67\% \pm 0.63$ &
 $95.09\% \pm 0.66$ &
 $95.21\% \pm 0.67$ &
 $95.51\% \pm 0.50$ &
 $95.34\% \pm 0.48$ &
 $\mathbf{96.03\% \pm 0.53}$
 \vspace*{0.0mm} \\
 \hline 

Avg rank
&5.20
&4.83
&3.67
&2.83
&2.93
&1.53
\\ \hline
    \end{tabular}
    \end{center}
\end{table*}

\subsection{Analysis of the results}

\subsubsection{Overall comparison}

A first general analysis that can be made by looking at Table \ref{tab:rescomp} is that the method $RFDis_{IH}$ is globally the best performing of the 6 methods compared. It is the method that has allowed to obtain the best average precision on 10 of the 15 databases, with an average rank of $1.53$. On the other hand, it comes as no surprise that the least efficient method here is the Euclidean distance method. As mentioned in the previous section, distance methods in general are very sensitive to high dimensions and many studies have shown that the Euclidean distance measure in particular can suffer from the curse of dimensionality \cite{Angiu2017, Feldb2019}. However, it is not the only explanations from our point of view: the Euclidean distance is the only one of the 6 dissimilarity measures that is not learnt from the data. The 5 other methods all compute their dissimilarity values taking into account the training instances.

The second analysis from a global point of view is that methods based on dissimilarities measured by Random Forests give better average accuracies than methods based on distances, including the LMNN method. The main reason is probably that this method, as most of the metric learning method, is based on the estimation of a large number of parameters, proportional to the number of features (most of the time equal to $m \times m$, $m$ being the number of features), and which therefore requires much more training instances. This is obviously not well suited for HDLSS problems and the use of a principal component analysis is not sufficient to overcome this. At the opposite, Random Forest methods are known for their robustness to high dimensions and also for their flexibility even with small sample size.

\begin{figure}[htbp]
    \centerline{\includegraphics[width=.95\linewidth]{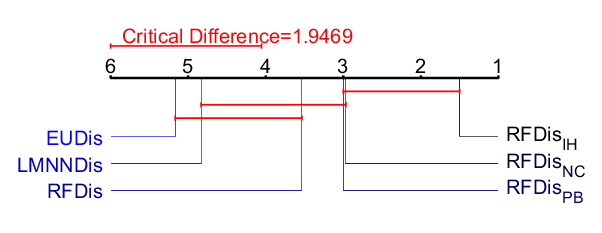}}
    \caption{The Nemenyi \textit{post hoc} test result for $\alpha$ = 0.05}
    \label{fig:cd}
\end{figure}

For a more rigorous comparison, we applied a statistical test of significance on these overall results. The test used in this experiment is the Nemenyi post-hoc test with Critical Differences (CD), as recommended in \cite{Demvs2006}. The result of this test is shown as a critical difference diagram in Figure \ref{fig:cd}. It allows to show that only the $RFDis_{IH}$ method is significantly superior to the distance-based methods and to the reference $RFDis$ method. Without taking into account the Euclidean distance, it can be seen that the mean rank difference between the other RF-based methods and the LMNN method is not statistically significant.

\subsubsection{Comparison of the $RFDis$ methods}

\begin{figure}[htbp]
    \centerline{\includegraphics[width=0.8\linewidth]{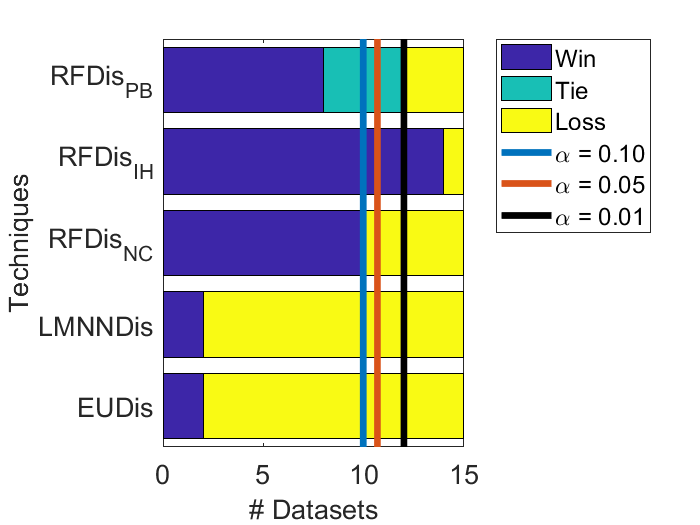}}
    \caption{Pairwise comparison between RFDis and all the other methods. The vertical lines illustrate the critical values considering a confidence level $\alpha = \{0.10, 0.05, 0.01\}$.}
    \label{3stat3} 
\end{figure}

The results are now presented by taking the $RFDis$ method as a baseline, in order to better highlight the improvement obtained by the proposed methods compared to this reference method. A pairwise analysis based on the Sign test is computed on the number of wins, ties and losses between $RFDis$ and all the other methods. The result, presented in Figure \ref{3stat3}, shows that the $RFDis$ method is a fairly solid baseline since most of the methods to which it is compared are not statistically better. Here again, the $RFDis_{IH}$ is the only one that is significantly better than the baseline for $\alpha = 0.10$, $0.05$ or even $0.01$. Nevertheless, the three $RFDis$ variants count more wins than losses against the reference $RFDis$, which tends to confirm our initial hypothesis that a finer RF dissimilarity measure leads to better results in this context.

On second reading, it can be noticed that our first proposal, the $RFDis_{NC}$ method, does not perform as well as $RFDis_{IH}$. The performance gaps with $RFDis$ are quite small in Table \ref{tab:rescomp} and the level of significance in Figure \ref{3stat3} is barely reached for $\alpha = 0.10$. It may be due to the following limitations of the proposed $RFDis_{NC}$ method: Firstly, as explained in Section \ref{ssec:rfdnc}, all the instances that lands in the same leaf node will share the same weight. However, this leaf may not be as reliable for estimating the dissimilarities of all these instances. For example, node \#10 in Figure \ref{fig:treeExplea} has many instances, some of which are close to the other nodes, while others are far away. Therefore, the reliability of this sub-region is not truly the same from one of its instances to another. The method does not take this phenomenon into account. Secondly, when the number of training instances is quite low, the number of OOB instances could be critically low, to the point that none of them are present in some terminal nodes. In this case, the posterior probability of the node is 1, but this does not mean that the corresponding sub-region is relevant for learning dissimilarity.

\section{Conclusion}

The Random Forest Dissimilarity (RFD) framework for multi-view learning is an efficient way to tackle multi-view classification tasks when in high dimensions and when very few instances are available for training (HDLSS problems for High Dimension, Low Sample Size). Such a situation is very common in the medical field for example, where data can be complicated to collect and complex to describe. The present work extends the RFD framework by proposing new methods for measure dissimilarities with Random Forest, that better suits to this specific learning context.

The goal of these proposals is to improve the standard RFD measure, that is based on rather coarse estimates given by the trees and essentially relies on averaging over a high number of trees in the forest. The tree measures proposed in this work are more accurate and better reflect the dissimilarities between instances with respect to the classification task, while remaining robust to high dimensions. The most efficient method is based on an instance hardness measurement calculated in the subspaces extracted from the trees of the RF. It allows to penalize unreliable dissimilarity estimates given by trees that have failed to correctly predict the instances. Experiments and results on real-world HDLSS multi-view datasets have shown that this mechanism is significantly more accurate than the standard RFD measure and than state-of-the-art metric learning methods. 

\section*{Acknowledgement}
\noindent This work is part of the DAISI project, co-financed by the European Union with the European Regional Development Fund (ERDF) and by the Normandy Region.

\bibliographystyle{IEEEtran}
\bibliography{ref}

\end{document}